**Agentic Explainable Artificial Intelligence (Agentic XAI) to Explore Better Explanations: A Case Study in Decision Support for Rice Cultivation in Japan**

Tomoaki Yamaguchi [a, b*], Yutong Zhou [c], Masahiro Ryo [c, d], and Keisuke Katsura [e]

[a] Faculty of Applied Biological Sciences, Gifu University, Yanagido, Gifu, 501-1193, Japan

[b] Center for Environmental and Societal Sustainability, Gifu University, Yanagido, Gifu, 501-1193, Japan

[c] Leibniz Centre for Agricultural Landscape Research (ZALF), Müncheberg, Germany

[d] Brandenburg University of Technology Cottbus–Senftenberg, Platz der Deutschen Einheit 1, Cottbus, Germany

[e] Graduate School of Agriculture, Kyoto University, Kitashirakawa-Oiwake-cho, Sakyo, Kyoto, 606-8502, Japan

E-mail addresses:

yamaguchi.tomoaki.t0@f.gifu-u.ac.jp (T.Y.);

Yutong.Zhou@zalf.de (Y.Z.);

Masahiro.Ryo@zalf.de (M.R.);

katsura.keisuke.2a@kyoto-u.ac.jp (K.K.);

[*]Correspondence: yamaguchi.tomoaki.t0@f.gifu-u.ac.jp (T.Y.); +81-58-293-2975 (T.Y.)

## Highlights

- A multimodal LLM iteratively refines SHAP output into farmer-facing advice
- 12 crop scientists scored 11 refinement rounds on 7 practitioner criteria
- Expert-judged quality peaked at Round 3 and then fell below the baseline
- 14 LLM judges overscore but preserve expert rank order (ρ = 0.58–0.90)
- Screened LLM judges can flag the quality peak to support stopping decisions

**Abstract**

Explainable artificial intelligence (XAI) reveals how explanatory variables relate to a response variable, yet communicating XAI outputs to laypersons remains difficult, limiting trust in AI-based predictions. Large language models (LLMs) can translate technical explanations into accessible narratives, but iterative refinement of XAI explanations by an autonomous LLM agent remains unexplored. This study proposes an agentic XAI framework that combines SHapley Additive exPlanations (SHAP) with iterative refinement by a multimodal LLM and tests it as an agricultural recommendation system on rice yield data from 28 fields in Japan. From a SHAP result, the agent explored additional analyses across 11 refinement rounds (Rounds 0–10). Crop scientists (n = 12) and LLM judges (n = 14) scored every round on seven criteria: Specificity, Clarity, Conciseness, Practicality, Contextual Relevance, Cost Consideration, and Crop Science Credibility. Both groups found that refinement raised the average score by 30–33% over Round 0, peaking at Rounds 3–4, after which quality declined, below the starting point for crop scientists. Refinement therefore requires strategic early stopping, which challenges assumptions of monotonic improvement. Criterion-level trajectories indicate a bias–variance trade-off. Early rounds lacked Specificity (bias), whereas excessive iteration eroded Conciseness and raised Cost Consideration through ungrounded economic reasoning (variance). The LLM judges overscored every criterion by 1.4–2.3 points but largely preserved the experts' ranking of rounds (Spearman $\rho = 0.58$–$0.90$), so screened LLM judges can flag the quality peak despite unreliable absolute scores. Trustworthy agentic XAI also needs expert-anchored screening of LLM judges and transparent, verifiable refinement records.

# 1. Introduction

Explainable artificial intelligence (XAI) has become a standard framework for AI-driven data analysis in science. By quantifying the importance of explanatory variables and their relationships with the target, through feature importance and dependence plots, XAI lets researchers move beyond predictive accuracy to understand which factors are associated with the response variable (Rajbongshi et al., 2026; Manafi Varkiani et al., 2025; Ryo, 2022; Molnar, 2020; Lundberg & Lee, 2017; Ribeiro et al., 2016). Yet communicating XAI outcomes to laypersons remains difficult, which hinders trust in AI-based predictions (Heider et al., 2026; Schiller et al., 2025; Lipton, 2018; Doshi-Velez & Kim, 2017). In agriculture in particular, this communication gap directly limits the trust that growers and agronomists place in data-driven decision support, and thus the adoption of otherwise capable AI tools (Edet & Chowdhury, 2026; Gardezi et al., 2023).

Large language models (LLMs) can, in contrast, foster science communication with laypersons (Galaz et al., 2025). LLMs can transform technical explanations into natural-language narratives, generate counterfactual scenarios, and hold interactive dialogues that align with users' knowledge structures (Chen et al., 2025; De Clercq et al., 2024). LLM-based advisory systems have proved practical in applied domains. For instance, an agricultural advisory system has served over 15,000 farmers with an approximately 75% success rate in providing actionable guidance (Singh et al., 2024). Evaluations of conversational LLMs in agricultural extension roles show that LLMs can provide high-quality, structured responses that often surpass those of human extension agents in consistency and comprehensiveness, although they need grounding in local data and validation protocols before they can be relied on for critical decisions (Ibrahim et al., 2024; Tzachor et al., 2022). Recent advisory systems address the need for grounding by retrieving agricultural knowledge before generating an answer (Sawant et al., 2026), which on a benchmark of agricultural questions reduced hallucinations and improved factual consistency (Ravindran et al., 2026). Generative AI use has also been linked to greater advisory effectiveness among extension agents (Gouroubera et al., 2026), and LLMs have been proposed as a means of translating the outputs of geospatial AI into actionable advice under human oversight (Gakhar & Singh, 2026).

Given this narrative capability, a few studies have combined LLMs with XAI to explain data-driven discoveries in plain language (Bilal et al., 2025). Key predictors found by a SHapley Additive exPlanations (SHAP) analysis, for instance, can be described in natural language. However, no previous study has had domain experts assess XAI–LLM integration, so its usefulness remains unknown. Nor is it clear how far the AI should elaborate its explanation of the XAI results. We argue that elaboration generally improves explanations but faces a trade-off between simplicity and comprehensiveness. A simple explanation of the system is easier for laypersons or domain experts to digest, but it may discard critical information. Conversely, an overly comprehensive description may blur important points. We therefore hypothesize that explanation comprehensiveness has an optimum, by analogy with the bias–variance trade-off in machine learning. A simplistic explanation gives the user biased information, whereas an overly extensive explanation overwhelms the user's capacity to digest it.

To test this hypothesis, we introduce an 'agentic XAI' approach. Among possible definitions, we define agentic XAI as an approach in which an AI agent searches for the best explanation of modeled patterns by iteratively refining a natural-language description. Agentic AI is an emerging technique in which a multimodal LLM operates as an autonomous agent capable of reasoning through multiple steps, retrieving external knowledge, and validating its outputs through feedback loops (Acharya et al., 2025; Hosseini et al., 2025; Sapkota et al., 2025). Recent implementations such as ReAct (Yao et al., 2023) and AutoGen (Wu et al., 2023) have shown that agentic architectures work across diverse domains, and frameworks for agriculture now organize agents into closed loops of perception, decision-making, and action (Davcev et al., 2026; Li et al., 2026). One agentic decision support system for smart agriculture has also applied SHAP and LIME to make its models transparent (Swati et al., 2026). To our knowledge, however, no study has had an agent iteratively refine its explanation of XAI results, or examined how the quality of that explanation changes as refinement proceeds. Our ambition is to enable smoother knowledge transfer from data science to domain experts and practitioners by providing explanations that are both technically rigorous and practically meaningful.

As a case study, we investigated agricultural systems, where complex interactions between soil properties, meteorological conditions, and management practices make it hard to translate AI outputs into actionable guidance for farmers (Holzinger et al., 2022; Gardezi et al., 2023). We generated recommendations through 11 refinement rounds (Rounds 0–10) and had them scored on seven quality metrics by crop scientists (n = 12) and by LLM-as-judge evaluators (Zheng et al., 2023) (n = 14). The first question was whether overall quality follows the inverted U-shaped pattern that the bias–variance analogy predicts as refinement progresses. The two evaluator panels also let us ask a second question with direct practical stakes, namely to what extent scalable LLM judges can reproduce expert judgment in absolute scores, in the ranking of rounds, and in the location of the quality peak.

Three findings emerge. First, refinement is not monotonic. Expert-judged quality rises by roughly one third to a peak at Rounds 3–4 and then deteriorates below the unrefined baseline, tracing the hypothesized inverted U. Second, the seven quality criteria follow distinct trajectories, in which five trace the inverted U while conciseness declines from the outset and cost consideration improves without grounding data, showing that the trade-off cannot be escaped by simple iteration. Third, LLM judges overscore all criteria yet largely preserve the experts' ranking of rounds, so a screened LLM judge can flag the quality peak even though its absolute scores cannot be trusted. Together, these findings yield evidence-based design principles for trustworthy agentic XAI systems, foremost among them disciplined early stopping.

## 2. Materials and methods

### 2.1 Proposed framework

The proposed agentic XAI framework comprises three interconnected components that enable iterative refinement of agricultural recommendations: (1) XAI analysis to elucidate yield drivers from a tabular dataset (Fig. 1a), (2) multimodal LLM (MLLM)-driven iterative refinement of recommendations for farmers (Fig. 1b), and (3) systematic evaluation across multiple metrics (Fig. 1c). In this architecture, the MLLM behaves as an agent, progressively improving its outputs through self-reflection, supplementary analyses, and iterative refinement.

All materials for reproducing the agentic XAI workflow are available on Zenodo: https://doi.org/10.5281/zenodo.17876329 (Yamaguchi et al., 2026). The repository includes the "Prompts" folder and "PythonCode.ipynb", which chronicles the code generation process from the initial baseline (Round 0) through all refinement stages (Rounds 1–10). Additionally, the "Figures" and "Recommendations" folders contain the complete sequence of visualization outputs and farmer recommendations, documenting the step-by-step expansion of the agent's analytical scope.

The theoretical foundation draws from cybernetic principles of feedback control adapted to knowledge translation (Holzinger et al., 2022), and conceptually applies the ReAct paradigm (Yao et al., 2023) wherein reasoning traces and analytical actions are interleaved to improve outputs. The framework implements a self-reflective mechanism inspired by recent advances in iterative refinement methodologies (Madaan et al., 2023; Wei et al., 2022). At each iteration, the MLLM analyzes its previous outputs, identifies analytical gaps, generates supplementary visualizations and statistical analyses through code generation, and produces enhanced recommendations. This approach exploits the reasoning capabilities shown by chain-of-thought prompting, while domain-specific constraints keep the outputs practically grounded (Pang et al., 2024).

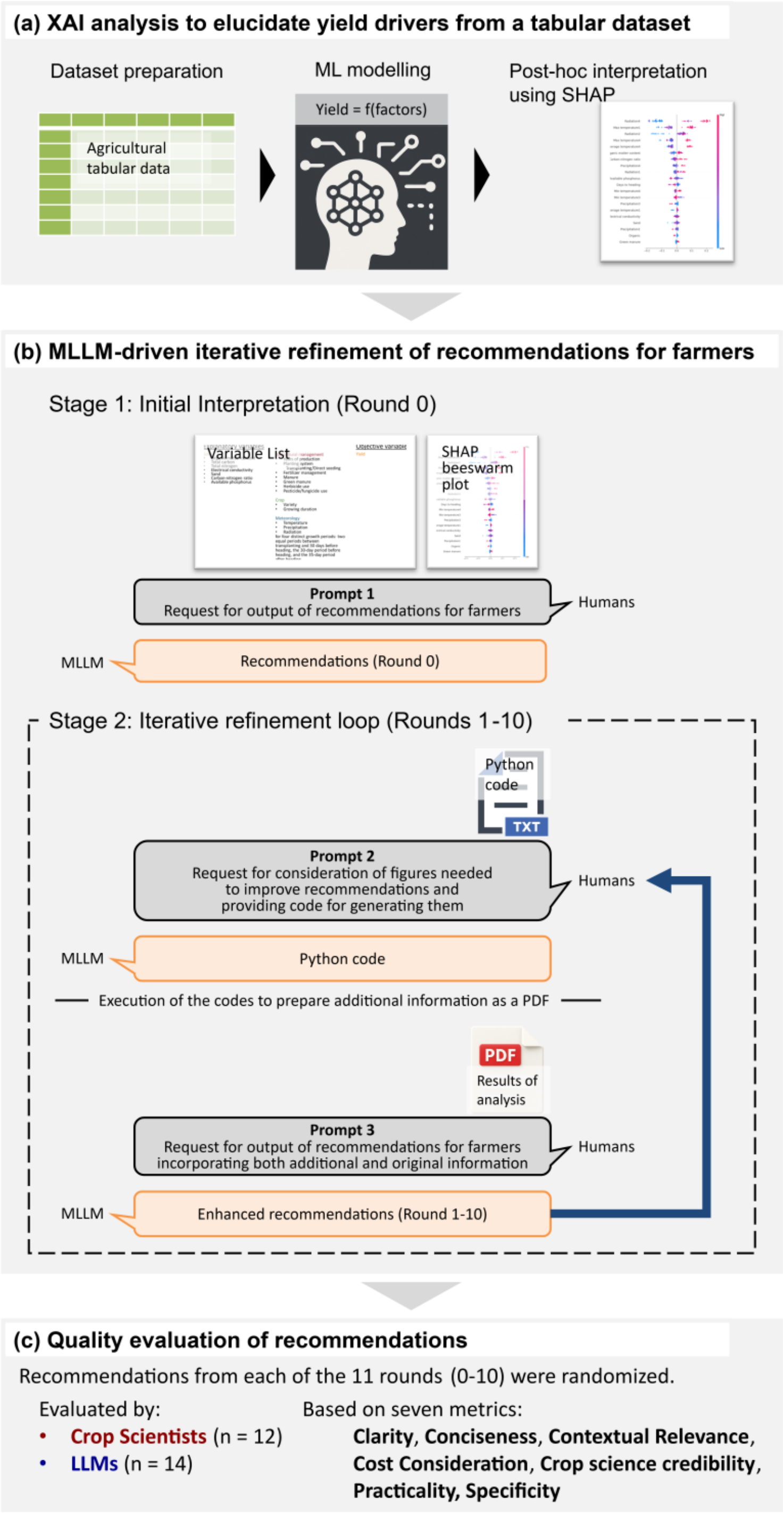

Fig. 1. Framework architecture of agentic explainable AI (XAI). The system integrates machine learning yield prediction with SHAP-based explanation and multimodal large language model (MLLM)-driven iterative refinement. (a) Input preparation proceeds from tabular data preparation, Random Forest modeling to SHAP visualization. (b) The agentic feedback loop enables iterative refinement where the MLLM analyzes XAI outputs, generates supplementary analytical code and visualizations, and produces progressively enhanced farmer recommendations through multiple refinement cycles. (c) Recommendations from all 11 rounds were randomized and evaluated by crop scientists (n = 12) and LLMs (n = 14) across seven metrics.

## 2.2 Dataset

We used a dataset collected from a Japanese farm that combined yield data with associated soil properties, rice varieties, cultivation management practices, and meteorological data. The dataset was collected in a single farm operation growing four rice varieties over three years (2021–2023) and comprises 28 fields, of which 16, 24, and 26 were observed in 2021, 2022, and 2023, respectively. The fields are located in Tomioka-machi, Futaba-gun, Fukushima Prefecture (37.34°N, 141.00°E) and cover approximately 7.5 ha in total (0.12–0.41 ha per field). We used four local rice varieties: Fukunoka (medium-height, used for sake production), Gohyakugawa (very early-ripening, with short stems), Sakurafukuhime (early-maturing, tall with strong stems), and Tennotsubu (medium-height, lodging-resistant).

Yield maps were generated from UAV imagery using a deep learning model (Xception) trained on maturity-stage images with three spectral bands (green, near-infrared, red edge), followed by spatial interpolation using the Inverse Distance Weighting method. The detailed methodology is described in Yamaguchi et al. (2025a). Soil properties were measured annually using a tractor-mounted real-time sensor system with hyperspectral capabilities (Kodaira and Shibusawa, 2013). Meteorological data were obtained from a gridded climatological database provided by the National Agriculture and Food Research Organization (NARO) (Ohno, 2014), with variables averaged across four growth periods: two equal periods between transplanting and 30 days before heading (early/late vegetative stage), the 30-day pre-heading period (reproductive stage), and the 35-day post-heading period (ripening stage).

All variables, including soil properties and yield, were averaged at the field level, resulting in 66 field-year observations across the three-year study period (n = 66). This dataset structure reflects typical smallholder farming contexts in East Asia, where field-level variation in soil properties, management practices, and environmental conditions creates complex yield-determining patterns that challenge conventional analytical approaches. The variables used in this study are summarized in Table 1, and their distributions are shown in Fig. S1.

Table 1. Description of variables used in the analysis, comprising the features (soil properties, agricultural management practices, and meteorological conditions) and the target variable (yield).

| Category | Variable | Unit | Description |
|---|---|---|---|
| Soil properties | Organic matter content | % | Soil organic matter content |
| | Total carbon | % | Total carbon content of the soil |
| | Total nitrogen | % | Total nitrogen content of the soil |
| | Electrical conductivity | dS $m^{-1}$ | Electrical conductivity of the soil |
| | Sand | % | Sand content of the soil |
| | Carbon-nitrogen ratio | – | Carbon-to-nitrogen ratio of the soil |
| | Available phosphorus | mg $P_2O_5$ (100 $g)^{-1}$ | Available phosphorus content of the soil |
| Agricultural management practices | Variety | – | Rice variety ID (Gohyakugawa (0), Fukunoka (1), Tennotsubu (2), Sakurafukuhime (3)) |
| | Days to heading | d | Number of days from transplanting (or seeding, for direct seeding) to heading |
| | Planting system | – | Direct seeding (0) or transplanting (1) |
| | Organic | – | Conventional cultivation (0), or organic or reduced-chemical cultivation with organic fertilizer (1) |
| | Green manure | – | No application (0) or application (1) |
| | Manure | – | No application (0) or application (1) |
| | Soil amendments | – | No application (0) or application (1) |
| | Chemical herbicide | – | No application (0) or application (1) |
| | Chemical pesticide and fungicide | – | No application (0) or application (1) |
| | Years of production | yr | Cumulative years of rice production in the field |
| Meteorological conditions | Average temperature1 | °C | Mean temperature for early vegetative stage |
| | Average temperature2 | °C | Mean temperature for late vegetative stage |
| | Average temperature3 | °C | Mean temperature for reproductive stage |
| | Average temperature4 | °C | Mean temperature for ripening stage |
| | Max temperature1 | °C | Mean daily maximum temperature for early vegetative stage |
| | Max temperature2 | °C | Mean daily maximum temperature for late vegetative stage |
| | Max temperature3 | °C | Mean daily maximum temperature for reproductive stage |
| | Max temperature4 | °C | Mean daily maximum temperature for ripening stage |
| | Min temperature1 | °C | Mean daily minimum temperature for early vegetative stage |
| | Min temperature2 | °C | Mean daily minimum temperature for late vegetative stage |
| | Min temperature3 | °C | Mean daily minimum temperature for reproductive stage |
| | Min temperature4 | °C | Mean daily minimum temperature for ripening stage |
| | Precipitation1 | mm $d^{-1}$ | Mean daily precipitation for early vegetative stage |
| | Precipitation2 | mm $d^{-1}$ | Mean daily precipitation for late vegetative stage |
| | Precipitation3 | mm $d^{-1}$ | Mean daily precipitation for reproductive stage |
| | Precipitation4 | mm $d^{-1}$ | Mean daily precipitation for ripening stage |
| | Radiation1 | MJ $m^{-2}$ $d^{-1}$ | Mean daily solar radiation for early vegetative stage |
| | Radiation2 | MJ $m^{-2}$ $d^{-1}$ | Mean daily solar radiation for late vegetative stage |
| | Radiation3 | MJ $m^{-2}$ $d^{-1}$ | Mean daily solar radiation for reproductive stage |
| | Radiation4 | MJ $m^{-2}$ $d^{-1}$ | Mean daily solar radiation for ripening stage |
| Yield | Yield | t $ha^{-1}$ | Rice yield |

### 2.3 XAI analysis to elucidate yield drivers from a tabular dataset

To establish the foundation for iterative refinement, we developed a Random Forest model for rice yield prediction from tabular agricultural data and applied SHAP analysis to generate explainability visualizations (Fig. 1a). These initial outputs served as the starting point for the LLM-driven iterative refinement process described in Section 2.4.

#### 2.3.1 Model development and configuration

All yield prediction models were implemented in Python 3.9.13 using scikit-learn 1.5.2, SHAP 0.46.0, NumPy 1.26.4, pandas 2.2.0, and matplotlib 3.8.2. The complete package environment is available in the accompanying repository, as described in the Data availability statement, to ensure reproducibility.

We employed a Random Forest regressor as the base model because of its robustness, interpretability, and proven effectiveness in agricultural applications (Raharimanana et al., 2023; Yamaguchi et al., 2025b). Model performance was evaluated using leave-one-out cross-validation (LOO-CV) across six hyperparameter configurations, systematically varying the number of estimators (100 or 200) and maximum tree depth (None, 10, or 20): (1) 100 estimators with unlimited depth, (2) 200 estimators with unlimited depth, (3) 100 estimators with a maximum depth of 10, (4) 200 estimators with a maximum depth of 10, (5) 100 estimators with a maximum depth of 20, and (6) 200 estimators with a maximum depth of 20. All models used a fixed random state of 42 as the seed for the pseudorandom number generator to ensure reproducibility. We selected the configuration with the highest LOO-CV $R^2$ and refitted it on the whole dataset as the final ML model.

The optimized Random Forest model, configured with 100 estimators and a maximum depth of 10, achieved an $R^2$ score of 0.749 in the LOO-CV setting. The $R^2$ score on the training dataset for the final ML model was 0.962 (Fig. S2). The gap between training and cross-validation performance indicates some overfitting, but the cross-validation score shows adequate predictive capability for generating SHAP-based explanations in this framework.

#### 2.3.2 SHAP-based interpretation

SHAP (TreeExplainer) was applied to the final ML model to generate feature importance scores and interaction effects (Lundberg & Lee, 2017). In agricultural contexts, this method has outperformed alternative XAI approaches, providing global and local interpretability at low computational cost (Ryo, 2022; Kaler & Kaur, 2025). We analyzed the top 20 most influential features based on mean absolute SHAP values. The primary visualization was a beeswarm plot showing the distribution of SHAP values for each feature, providing insights into both feature importance and directional effects on yield prediction.

SHAP analysis identified management timing variables, meteorological conditions during critical growth phases, and soil nutrient parameters as primary determinants of yield (Fig. 2). This SHAP beeswarm plot served as the input for the subsequent agentic XAI process, wherein the MLLM iteratively analyzed and refined cultivation management recommendations across multiple refinement rounds (Rounds 0–10).

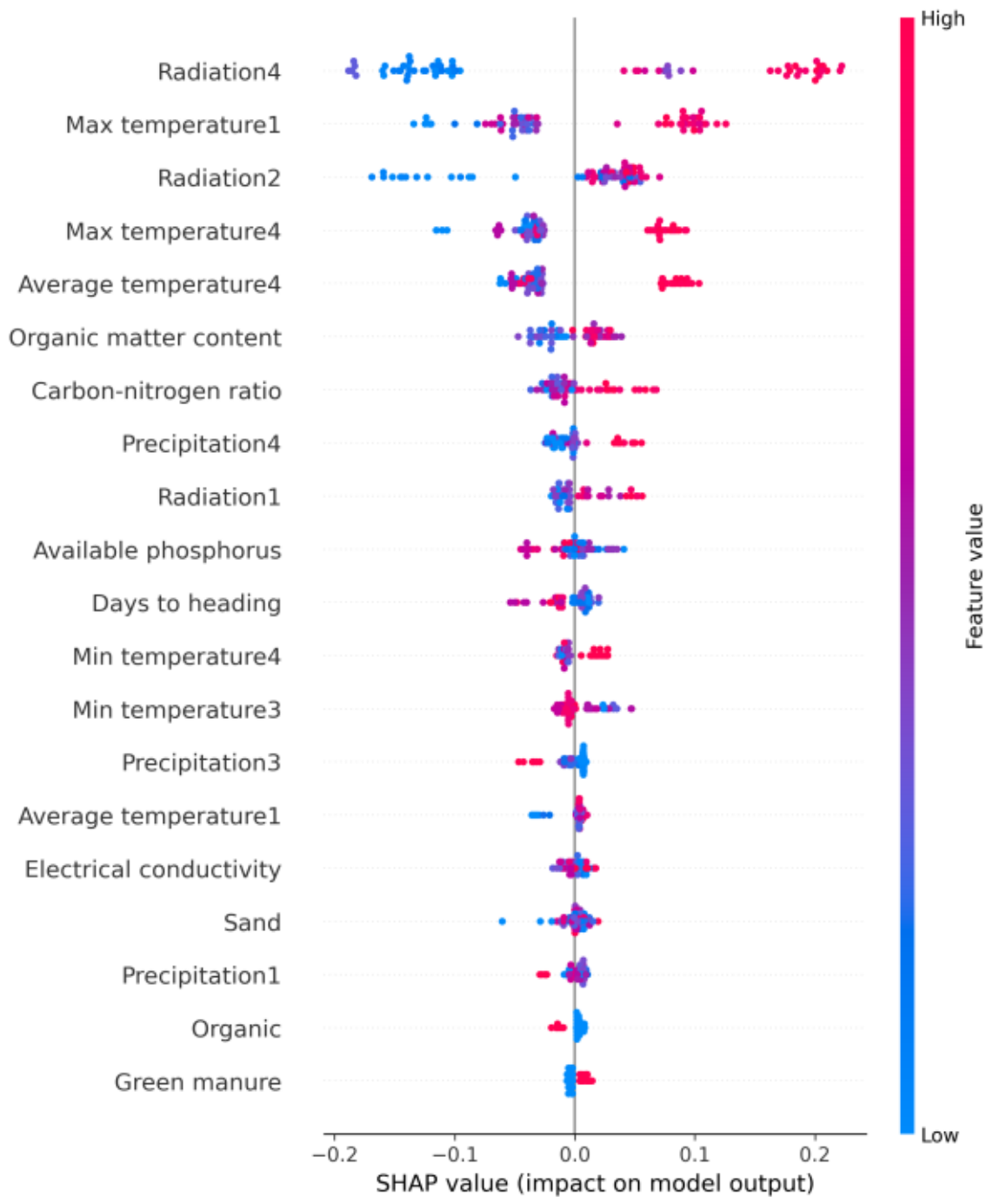

Fig. 2**.** SHAP beeswarm plot of the Random Forest yield model, which served as the initial input for the agentic XAI refinement process. Each point represents a field-year observation (n = 66), with its position along the x-axis indicating the SHAP value (impact on model output, in t $ha^{-1}$) and its color the feature value (red, high; blue, low). The 20 features with the largest mean absolute SHAP values are shown in descending order. The number suffix (1–4) of a variable denotes the growth period, where 1 and 2 are two equal periods between transplanting and 30 days before heading, 3 is the 30-day pre-heading period, and 4 is the 35-day post-heading period.

## 2.4 MLLM-driven iterative refinement of recommendations for farmers

The agentic XAI framework uses a multimodal LLM to run an iterative analysis loop (Fig. 1b) that implements principles from recent work on self-refinement and iterative reasoning optimization (Madaan et al., 2023; Kumar et al., 2024). We employed Claude Sonnet 4 (Anthropic) with standard settings: web search disabled, normal response style, and extended thinking disabled. In this sequential workflow, the MLLM alternates between interpreting data, generating Python code that extracts new statistical evidence, and updating its recommendations on that evidence.

The workflow consists of an initialization phase followed by a defined number of refinement iterations (Rounds 1–10). The specific inputs and prompts for each stage are detailed below, with the three prompts reproduced in full in Text S1:

**Stage 1: Initial Interpretation (Round 0)**

- **Input:** The MLLM is provided with the SHAP beeswarm plot image and a list of variables.
- **Prompt 1:** Request for output of recommendations for farmers.
- **Action:** The model generates a baseline textual interpretation and set of recommendations based solely on the visual features of the SHAP plot.

**Stage 2: Iterative Refinement Loop (Rounds 1–10)**

In each subsequent round, the framework executes a three-step cycle to update the recommendations:

1. **Gap Analysis and Python Code Generation:**
   - **Prompt 2:** Request for consideration of figures needed to improve recommendations and providing code for generating them
   - **Action:** The MLLM reviews its previous output and the dataset structure to identify variables that require quantitative verification (e.g., specific correlation coefficients or feature interactions not visible in the initial plot). The model then outputs executable Python code intended to calculate these statistics or generate supplementary plots.
2. **Execution and Data Retrieval:**
   - **Action:** The generated Python code is executed on the tabular dataset. The resulting outputs (e.g., statistical summaries, correlation matrices, or temporal interaction plots) are compiled into a PDF document.
3. **Synthesis and Update:**
   - **Prompt 3:** Request for output of recommendations for farmers incorporating both additional and original information.
   - **Action:** The MLLM receives the new PDF document alongside the original SHAP plot. It synthesizes the quantitative results from the PDF with the visual data to rewrite the recommendations.

**Context Management:** Throughout the process, the conversation history is preserved. This ensures that in Round N, the model generates responses conditional on the accumulated information from Rounds 0 through N − 1, allowing the specific details in the recommendations to become more granular as the model accesses more calculated statistics.

## 2.5 Quality evaluation of recommendations

### 2.5.1 Evaluation design and protocol

Because agricultural advisory quality is hard to evaluate, we used two evaluator groups, human experts (crop scientists) and AI systems (LLMs) (Fig. 1c). Twelve crop scientists evaluated the recommendations so that domain knowledge, including Japanese farming conditions, practical constraints, and established agronomic principles, entered the evaluation. These crop scientists were PhD holders in agricultural sciences and current or former members of the Crop Science Society of Japan. Participation was voluntary, and consent was obtained after the crop scientists had been informed of the purpose of the study and of how their evaluation data would be used. The evaluations were recorded and analyzed anonymously. This human expert evaluation was complemented by assessments from 14 LLMs representing five major AI providers: OpenAI (ChatGPT 4o, ChatGPT 5.1 Instant, ChatGPT 5.1 Thinking), Anthropic (Claude Sonnet 4, Claude Opus 4.1, Claude Sonnet 4.5), xAI (Grok 3 Fast, Grok 3 Expert, Grok 4.1), DeepSeek (DeepSeek, DeepSeek with DeepThink), and Google (Gemini 2.5 Flash, Gemini 2.5 Pro, Gemini 3.0 Pro). Recent studies have demonstrated that properly implemented LLM evaluators can achieve 75–85% agreement with human expert judgments (LLM-as-judge), particularly when using multiple models to mitigate individual biases (Zheng et al., 2023; Liu et al., 2023; Kim et al., 2024a, 2024b).

Every evaluator, crop scientist or LLM, independently scored the recommendations of all 11 rounds (Rounds 0–10) based on seven evaluation metrics. Drawing on agricultural decision support system evaluation frameworks that emphasize accessibility, usability, and practical implementation (Zhai et al., 2020), user acceptance and workload considerations (Htun et al., 2022), and economic value assessment (Helps et al., 2024), we developed the seven evaluation metrics (Table 2), with scores ranging from 1 for very poor to 7 for excellent. Although each criterion in Table 2 is described by several guiding questions, evaluators weighed these aspects holistically and assigned a single integer rating per criterion, so that each round received exactly seven criterion-level scores from every evaluator.

Table 2. Evaluation metrics for assessing recommendation quality.

| Metrics | Definition |
|---|---|
| Clarity | Is the response easy to understand, well-structured, and free from ambiguity? Is the logical flow clear? Is technical terminology explained when necessary? Can farmers grasp the main points without confusion? |
| Conciseness | Is the response succinct and to the point without unnecessary verbosity? Does it avoid redundant information or excessive repetition while still being complete? Can busy farmers read and understand it quickly during their work breaks? |
| Contextual Relevance | Does it avoid over-generalization or excessive theoretical discussion that doesn't fit the given background? Does the response align with the actual conditions and context provided in the dataset (location, varieties, field scale, management practices)? Is the advice appropriate for the specific farming scenario described? |
| Cost Consideration | Does the response address economic factors such as costs, budget requirements, return on investment, or cost-effectiveness? Does it consider the financial constraints that farmers typically face? |
| Crop Science Credibility | Is the response agronomically sound and reliable from a crop science perspective? Does it reflect established scientific principles in agronomy, plant physiology, soil science, and crop management? Is the information accurate and credible based on current agricultural research and best practices? |
| Practicality | Does the response offer actionable and feasible advice or solutions that can be realistically implemented? Are the suggestions achievable with typical farm resources, skills, and time constraints? |
| Specificity | Does the response provide detailed and precise information rather than being vague or general? Does it include specific numbers, measurements, varieties, methods, or step-by-step instructions that farmers can directly reference? |

Before the evaluation, the crop scientists received background information in writing and orally in an online meeting, and the LLMs received the same information in writing. The instructions, the criteria as they were presented and this background are reproduced in Text S2. This background described the dataset context (Section 2.2) and the 37 input variables categorized into soil properties (7 variables), agricultural management practices (10 variables), and meteorological conditions (20 variables across four growth stages), with yield as the target variable (Table 1). Evaluators were told that the recommendations came from an iterative process in which the LLM analyzed the tabular data and revised its outputs over 11 cycles (Rounds 0–10), and that each response contained agronomic insights into the relationships between explanatory variables and rice yield together with practical recommendations for local farmers.

To minimize systematic biases such as position effects and verbosity bias documented in LLM evaluation studies (Ye et al., 2024; Chern et al., 2024), we applied the same evaluation protocol to both crop scientists and LLMs. For each evaluator, the sequence in which the 11 recommendations (Rounds 0–10) were presented was randomized, and the order used is given in Text S2. Each evaluator provided independent assessments without knowledge of other evaluators' scores or the iteration number being evaluated, ensuring blind evaluation.

To keep the LLM evaluations independent of one another, all LLM evaluations were conducted in isolated conversation contexts. For ChatGPT, Claude, and Grok models, evaluations were performed in incognito/private mode to prevent any influence from previous conversation history. For DeepSeek and Gemini models, which do not offer an incognito mode, evaluations were conducted in completely new conversation threads with no prior conversation history.

### 2.5.2 Statistical analysis of quality trajectories

After collecting the assessments, we organized the evaluation data separately for crop scientist evaluations ($n = 12$) and LLM evaluations ($n = 14$), treating each evaluator (or repetition) as an independent replicate within its group. For each round (0–10), we computed an overall quality score for each evaluator by averaging the seven evaluation metrics. For each metric and the overall score, we then calculated descriptive statistics across evaluators within each group for each round (mean and standard error of the mean, SEM).

To test whether recommendation quality differed across rounds, we conducted one-way ANOVA for each metric and the overall score. The null hypothesis assumed equal means across rounds, and statistical significance was evaluated at $\alpha = 0.05$.

To assess whether recommendation quality exhibited an inverted U-shaped trajectory across rounds, we fitted a generalized additive model (GAM) relating round number (X) to the round-level mean score (Y):

$$Y = \beta_0 + f(X) + \varepsilon$$

where f(X) represents a smooth function estimated via penalized regression splines with smoothing parameters selected by internal grid search.

Detection of an inverted U-shape was based on the first derivative of the fitted smooth, following derivative-based change detection approaches for GAMs (Simpson, 2018; von Brömssen et al., 2021). The derivative was evaluated by numerical differentiation over a dense grid of round values. We classified a trajectory as inverted U-shaped when all of the following criteria were met: (i) exactly one zero-crossing occurred within the observed range at which the derivative changed sign from positive to negative; (ii) the derivative was positive on average in early rounds and negative on average in later rounds; and (iii) the 95% confidence interval for the derivative excluded zero in both regions, indicating a statistically robust increase followed by a robust decrease. The peak round was defined as the zero-crossing location, and the peak score was obtained from the fitted GAM curve at that round.

In our implementation, 95% confidence intervals for the GAM mean function were computed as pointwise intervals from the fitted penalized spline model, using the estimated coefficient covariance matrix derived from the (penalized) design matrix and the residual variance. Confidence intervals for the first derivative were then approximated by numerically differentiating the lower and upper 95% confidence interval curves for the mean function across the evaluation grid.

Additionally, to evaluate whether the nonlinear GAM provided a better description of the observed trajectories than a simple linear trend, we fitted a linear ordinary least squares regression model as a baseline:

$$Y = \alpha_0 + \alpha_1 X + \varepsilon$$

Model fit was summarized using the coefficient of determination ($R^2$). To formally compare model performance while accounting for model complexity, we computed Akaike's Information Criterion (AIC) (Akaike, 1973) under a Gaussian error assumption for both the GAM and the linear model:

$$AIC = n[\log(2\pi) + 1 + \log(RSS/n)] + 2k$$

where n denotes the number of observations (rounds), RSS represents the residual sum of squares, and k indicates the number of estimated parameters (effective degrees of freedom for GAM; $k = 2$ for the linear model).

Following the widely cited guidelines of Burnham and Anderson (2002), models with $\Delta AIC < 2$ are considered to have substantial empirical support. We required the GAM to outperform the linear model by at least 2 AIC units ($\Delta AIC = AIC_{Linear} - AIC_{GAM} > 2$) to justify adopting the more complex nonlinear model. The GAM was therefore adopted only when it both satisfied the derivative-based criteria for an inverted U-shaped pattern and demonstrated substantially improved model fit relative to the linear baseline. When this criterion was not met, the linear model was considered sufficient to describe the trend and was used as the reference representation. This comparison-based procedure ensures that nonlinear structure was adopted only when strongly supported by the data, thereby reducing the risk of overfitting and avoiding circular inference regarding peak detection.

All analyses were conducted in Python 3.12.12 within the Google Colab environment. Data processing and statistical analyses were performed using NumPy (v2.0.2), SciPy (v1.16.3), and Pandas (v2.2.2), with visualization implemented in Matplotlib (v3.10.0). Generalized additive models were estimated using pyGAM (v0.12.0). The analysis code archived in the Zenodo record re-runs every analysis reported here, including the yield model of Section 2.3.1, in a single pinned environment (Python 3.11.9) and reproduces the reported values.

### 2.5.3 LLM–expert agreement analysis

To quantify whether, and at which level of measurement, the LLM evaluators can stand in for the crop scientists, we compared the two evaluator groups on each of the eight score series (seven criteria plus their average), using the round-level group means ($n = 11$ rounds). Absolute agreement was assessed by the mean score difference (LLM − crop scientists) with paired t tests and Cohen's d, and by Lin's concordance correlation coefficient (CCC) (Lin, 1989) with bootstrap 95% confidence intervals; Pearson's r is reported alongside the CCC to separate linear association from absolute concordance. Rank agreement, defined as whether the two groups order the 11 rounds consistently, was assessed by Spearman's rank correlation ($\rho$) of the round means, with block-bootstrap 95% confidence intervals. Peak agreement was assessed by comparing the round at which each group's observed mean score reached its maximum.

Because a deployed advisory system would more realistically rely on one or a few LLM judges than on a 14-model panel, we additionally benchmarked each LLM individually. For every LLM and criterion, we computed the absolute difference between that LLM's peak round and the crop-scientist consensus peak round, and summarized each LLM by the proportion of criteria for which its peak lay within ±1 round of the expert consensus.

## 3. Results

### 3.1 Overall quality dynamics across refinement rounds

The overall average score (mean of seven evaluation metrics) showed a consistent inverted U-shaped pattern across the 11 refinement rounds (Rounds 0–10) in both evaluator groups, with peak performance occurring in early-to-middle iterations followed by substantial deterioration (Fig. 3, Tables S1 and S2).

One-way ANOVA indicated significant differences across rounds ($p < 0.001$), and GAM analyses detected an inverted U-shaped trajectory for the average score in both crop scientists and LLMs (Table S2), indicating improvement in the early-to-mid rounds followed by decline. In crop scientist evaluations, the observed average score increased from 3.68 at Round 0 to a maximum of 4.91 at Round 3 ($\Delta = +1.23$), then declined to 2.64 by Round 10 ($\Delta = -2.26$ from the observed peak) (Fig. 3a; Table S1). The GAM also supported an inverted U-shaped trajectory with an estimated peak at Round 2.50 (estimated peak value 4.523) and better fit than the linear baseline ($\Delta AIC = 9.25$) (Table S2). In LLM evaluations, the observed average score increased from 4.78 at Round 0 to a maximum of 6.21 at Round 4 ($\Delta = +1.44$), then decreased to 5.18 by Round 10 ($\Delta = -1.03$ from the observed peak) (Fig. 3b; Table S1). The GAM likewise detected an inverted U-shaped trajectory with an estimated peak at Round 3.53 (estimated peak value 6.028) and strong support relative to the linear model ($\Delta AIC = 13.42$) (Table S2).

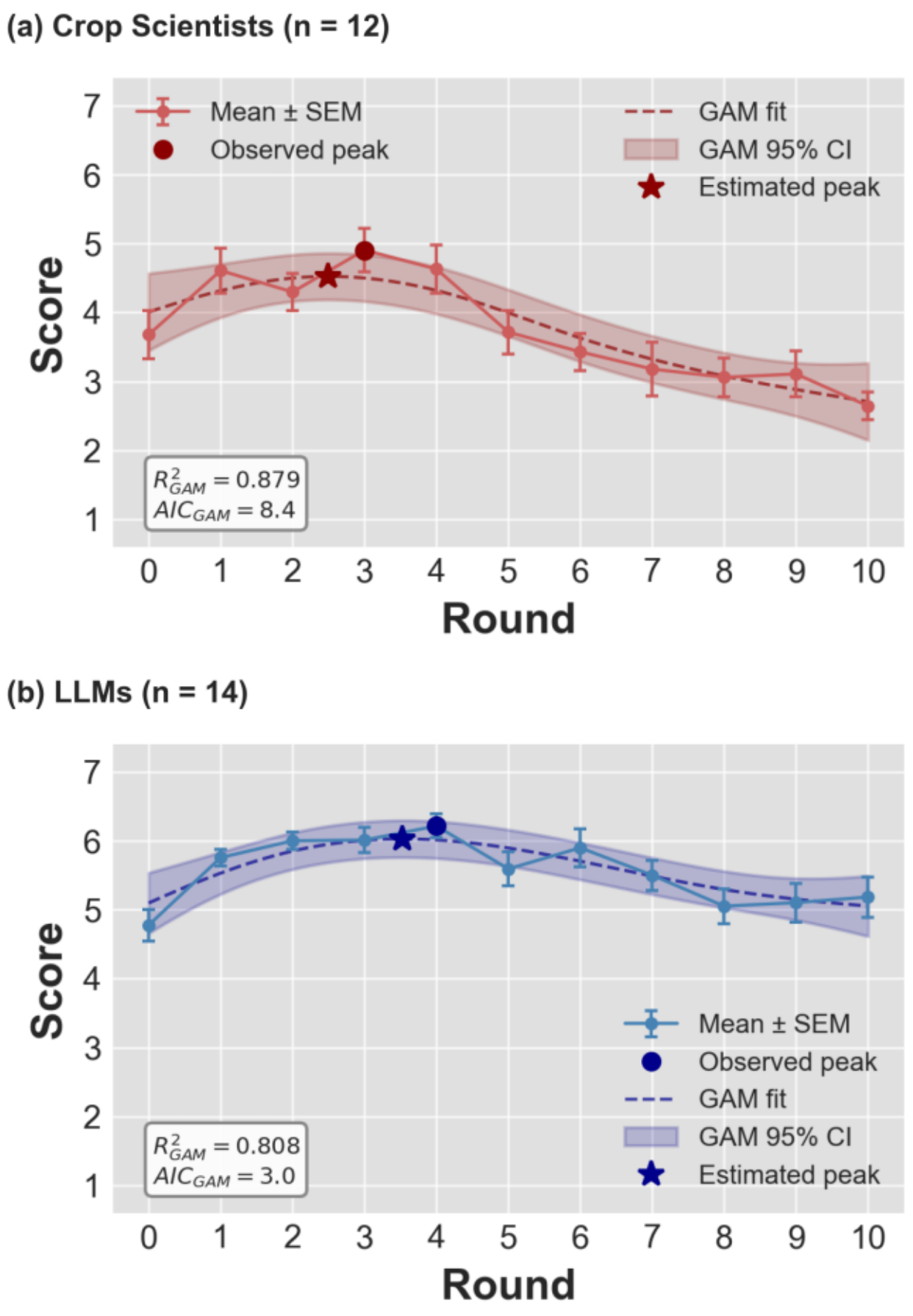

Fig. 3**.** Average scores (mean of seven evaluation metrics) across rounds for crop scientists (a) and LLMs (b). Points show means and error bars indicate the standard error of the mean (SEM). Dark markers indicate the observed peak (maximum mean value). GAM fit is shown as a dashed curve with shaded 95% confidence intervals; stars mark the estimated peak (derivative zero-crossing).

The score rise is visible in the advice itself. Fig. 4 sets side by side the Round 0 and the Round 3 excerpts on one advisory topic that both rounds addressed, early-season temperature management. Round 0 identified the relevant levers but left them unquantified, recommending varieties of appropriate maturity and shallow flooding during hot periods. Round 3 restated the same levers as monitorable thresholds with an action attached to each, a daily maximum temperature above 25 °C for the first 40 days after transplanting, a water depth of 3–5 cm during cool periods and not below 2 cm, and protective measures once maximum temperatures stayed below 22 °C for five consecutive days. Consistent with this, Specificity gained more between Round 0 and its observed peak than any other criterion in the crop-scientist panel (+2.17), and more than any criterion except Cost Consideration in the LLM panel (+2.93; Table S1).

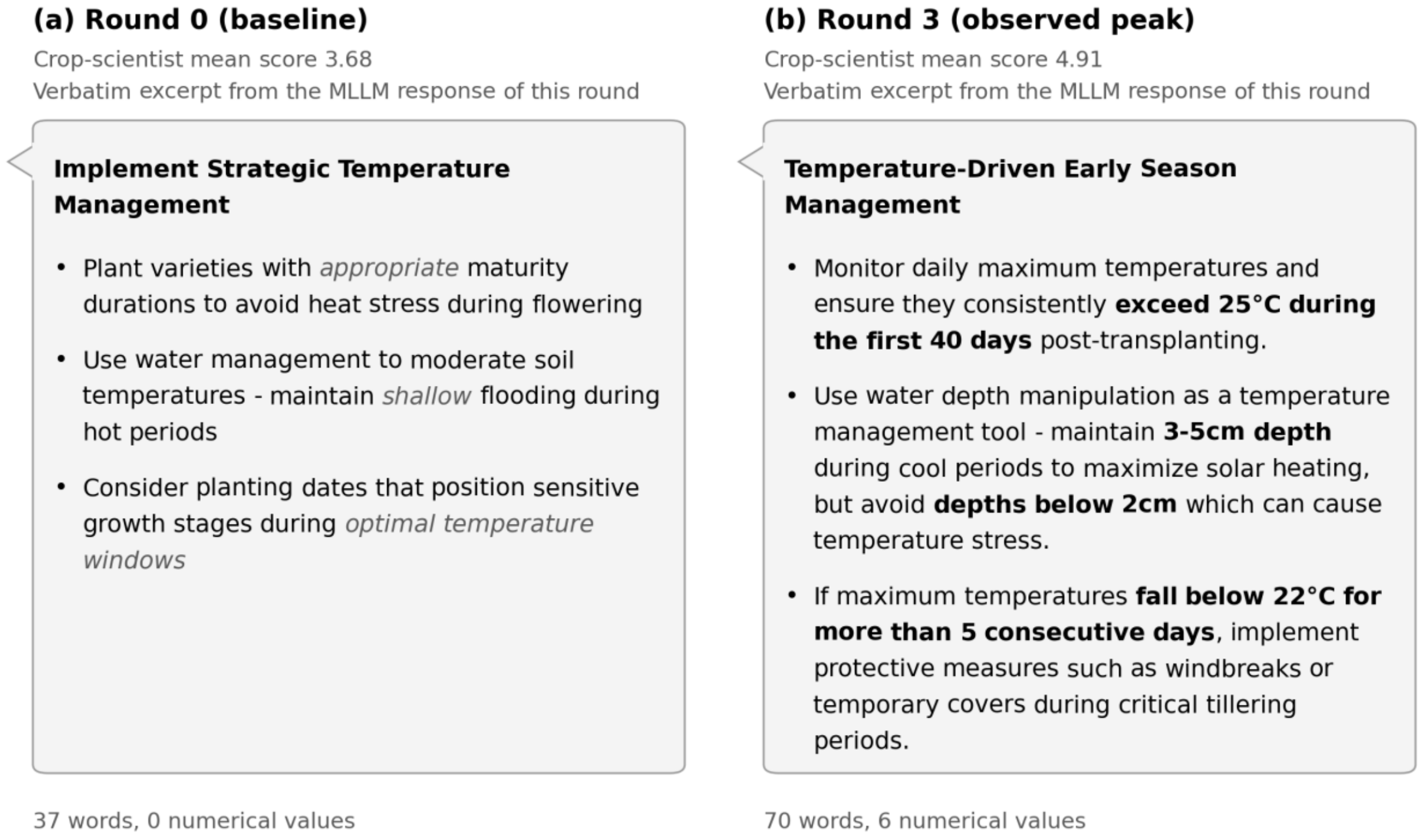

Fig. 4. Excerpts from the MLLM responses in Round 0 (a) and Round 3 (b). Each box quotes verbatim the section of the response on early-season temperature management; the rest of each response is omitted, and the full responses of all 11 rounds are deposited on Zenodo. Italics in (a) mark the qualifiers that were left undefined, and bold in (b) marks the quantitative content that replaced them. The counts below each box refer to the text quoted under each heading. Round 3 is the observed peak of the crop-scientist average (Fig. 3a).

## 3.2 Metric-specific performance dynamics

Across metrics, the observed trajectories (Table S1) and GAM-based trajectory classification (Fig. 5; Table S2) indicated three distinct patterns: (i) inverted U-shaped dynamics (background gray) for five metrics, (ii) monotonic increases under linear trends (background orange) for Cost Consideration, and (iii) monotonic decreases under linear trends (background blue) for Conciseness.

Five metrics (Specificity, Clarity, Practicality, Contextual Relevance, and Crop Science Credibility) showed inverted U-shaped dynamics in both evaluator groups. These metrics exhibited early-to-mid observed peaks (typically Rounds 2–4) in both crop scientists and LLMs (Table S1). GAM analyses detected inverted U-shaped patterns for these five metrics in the crop scientists, and for four of the five in the LLMs, with improved fit relative to linear baselines ($\Delta AIC > 2$) (Table S2), supporting an early improvement followed by later decline. The exception was Clarity in the LLM panel, where scores stayed uniformly high and no significant trend was detected.

Cost Consideration instead improved steadily throughout the rounds, with late-round observed peaks. For crop scientists, the observed peak occurred at Rounds 7–8 (4.667), with Round 10 maintaining a high value (4.500) (Table S1). Similarly, LLM evaluations showed the observed peak at Round 9 (6.929), with Round 10 remaining near this peak level (6.714) (Table S1). Accordingly, the GAM criteria did not identify an inverted U-shaped pattern, and linear regression showed significant positive slopes for Cost Consideration in both evaluator groups (Fig. 5; Table S2), consistent with monotonic improvement.

In contrast, Conciseness showed early peak performance followed by consistent decline. Crop scientists recorded observed peaks at Rounds 1 and 3 (4.833), which subsequently declined to 2.333 by Round 10 (Table S1). LLM evaluations showed an even more pronounced pattern, with peak performance occurring at Round 0 (5.929) before declining to 5.143 by Round 10 (Table S1). In line with these observations, GAM criteria did not detect an inverted U-shaped pattern, and linear regression showed significant negative slopes for Conciseness in both evaluator groups (Fig. 5; Table S2), indicating progressive deterioration throughout the evaluation period.

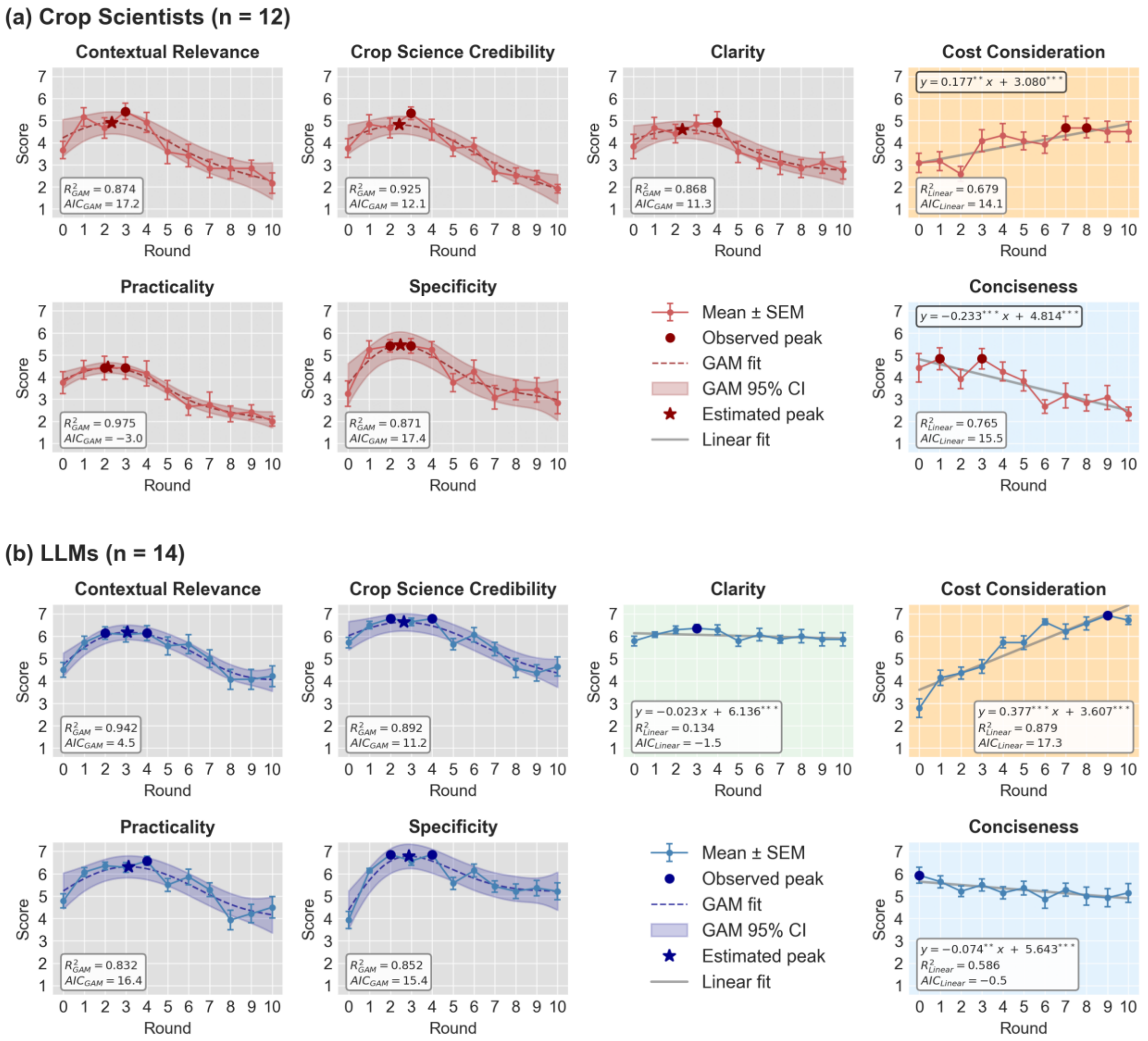

Fig. 5. Individual evaluation metric scores across rounds for crop scientists (a) and LLMs (b). Seven metrics are shown: Clarity, Contextual Relevance, Crop Science Credibility, Practicality, Specificity, Conciseness, and Cost Consideration. Points represent mean scores, and error bars indicate the standard error of the mean (SEM). Dark markers indicate observed peaks (maximum mean values). When an inverted U-shaped relationship was identified using derivative-based criteria and the generalized additive model (GAM) provided a better fit than the linear model ($\Delta AIC > 2$), the GAM fit is displayed as a dashed curve with shaded 95% confidence intervals; stars mark the estimated peak (derivative zero-crossing). Otherwise, linear regression is shown as a solid gray line with the fitted slope and intercept annotated; asterisks indicate statistical significance (* $p < 0.05$, ** $p < 0.01$, *** $p < 0.001$). Background shading indicates the inferred trend type: gray for inverted U-shaped patterns, green for no trend (no linear slope), orange for monotonic increases (positive linear slope), and blue for monotonic decreases (negative linear slope).

### 3.3 Progression of analytical complexity across refinement rounds

The agent's analytical scope expanded steadily across the 11 refinement rounds (Table 3). New statistical methods and visualization types appeared in almost every round, and the cumulative number of figures grew from 1 to 93.

The analytical focus also shifted. Early rounds (0–3) established mechanistic understanding of yield-limiting factors, management strategies, and environmental interactions. Middle rounds (4–8) introduced economic analysis, field typology classification, and sustainability assessment. Later rounds (9–10) adopted a critical stance, questioning whether increased analytical complexity improved farmer outcomes, while synthesizing recommendations with risk warnings.

Table 3. Progression of analytical methods and visualizations across 11 refinement rounds. New Statistical Methods and New Visualization Types show contents first introduced in each round. Number of Figures indicates figures generated in each round (cumulative total in parentheses).

| Round | New Statistical Methods | New Visualization Types | Number of Figures | Analytical Focus |
|---|---|---|---|---|
| 0 | Random forest, SHAP | Beeswarm plot | 1 | Identify key yield-limiting factors using explainable AI |
| 1 | Correlation matrix | Bar chart, Heatmap, Line plot, Scatter plot, Violin plot | 8 (9) | Compare management strategies and identify critical growth stages for intervention |
| 2 | PCA, Q-Q plot, Explained variance | Histogram, Pie chart | 17 (26) | Understand weather pattern effects and soil-yield relationships for site-specific management |
| 3 | Response curve (binning analysis) | Error bar plot, Text annotation | 10 (36) | Quantify environmental factor interactions and provide management timeline |
| 4 | Partial dependence, Return on investment | Box plot, Flowchart | 9 (45) | Assess economic feasibility and return on investment for precision management adoption |
| 5 | K-means clustering, Response surface methodology, Sensitivity analysis | Contour plot, Roadmap | 9 (54) | Identify field typologies and target management strategies by yield potential class |
| 6 | Decision tree, Spearman's rank correlation | Filled area plot, Boxed note | 7 (61) | Evaluate long-term profitability and sustainability under climate/market uncertainty |
| 7 | Kolmogorov-Smirnov test, Polynomial regression | - | 10 (71) | Determine critical thresholds for intervention and quantify implementation risks |
| 8 | Monte Carlo simulation, DBSCAN clustering, Value at risk | Bubble chart | 6 (77) | Assess regional transferability and farmer-specific constraints for technology adoption |
| 9 | Anderson-Darling test | Radar plot | 8 (85) | Evaluate whether increased analytical complexity actually improves farmer outcomes |
| 10 | Beta distribution simulation | Stacked bar chart | 8 (93) | Synthesize actionable recommendations with clear risk warnings and exit criteria |

### 3.4 LLM–expert agreement

Table 4 summarizes the agreement between the two evaluator groups. At the level of absolute scores, the LLM panel rated every criterion higher than the crop scientists, by 1.42–2.27 points on the 7-point scale (paired t tests, all $p < 0.001$; Cohen's $d = 1.76$–$4.97$). Lin's CCC was correspondingly low (0.04–0.38) even though Pearson's r reached 0.72–0.94, indicating a strong linear association displaced by a large constant shift.

Table 4. Agreement between LLM and crop-scientist evaluations across the eight score series (seven criteria and their average). Mean diff is the difference in grand means (LLM − crop scientists) on the 7-point scale; Cohen's d accompanies paired t tests (all $p < 0.001$). Lin's CCC and Spearman's ρ are computed on round-level group means ($n = 11$ rounds), with bootstrap 95% confidence intervals. Peak round reports each group's observed peak round or rounds, with ties listed (crop scientists / LLMs). CS, crop scientists.

| Metric | Mean diff (LLM − CS) | Cohen's d | Pearson r | Lin's CCC [95% CI] | Spearman ρ [95% CI] | Peak round (CS / LLM) |
|---|---|---|---|---|---|---|
| Average | +1.80 | 3.40 | 0.72 | 0.12 [0.01, 0.18] | 0.76 [0.09, 0.94] | 3 / 4 |
| Clarity | +2.27 | 3.31 | 0.74 | 0.04 [−0.00, 0.06] | 0.58 [−0.28, 0.79] | 4 / 3 |
| Contextual Relevance | +1.42 | 2.33 | 0.84 | 0.38 [0.09, 0.55] | 0.81 [0.34, 0.88] | 3 / 2, 4 |
| Crop Science Credibility | +2.08 | 4.97 | 0.94 | 0.28 [0.10, 0.36] | 0.90 [0.37, 0.99] | 3 / 2, 4 |
| Practicality | +2.06 | 3.59 | 0.81 | 0.22 [0.05, 0.32] | 0.83 [0.39, 0.94] | 2, 3 / 4 |
| Specificity | +1.63 | 2.97 | 0.84 | 0.32 [0.07, 0.48] | 0.88 [0.43, 0.95] | 2, 3 / 2, 4 |
| Conciseness | +1.62 | 2.40 | 0.75 | 0.11 [0.03, 0.15] | 0.80 [0.45, 0.90] | 1, 3 / 0 |
| Cost Consideration | +1.53 | 1.76 | 0.81 | 0.32 [0.01, 0.45] | 0.71 [−0.19, 0.92] | 7, 8 / 9 |

Rank-level agreement, by contrast, was substantial. Spearman's ρ between the two groups' round-mean rankings ranged from 0.58 (Clarity) to 0.90 (Crop Science Credibility), and the two groups' observed peaks agreed exactly on one score series and lay within one round of each other on all eight (Table 4).

The panel average, however, conceals large differences among individual judges (Fig. 6). The best-performing judge (Gemini 3.0 Pro) located the expert consensus peak within ±1 round on seven of the eight score series (87.5%), whereas the weakest (Claude Sonnet 4), the same model that generated the recommendations, did so on only one (12.5%).

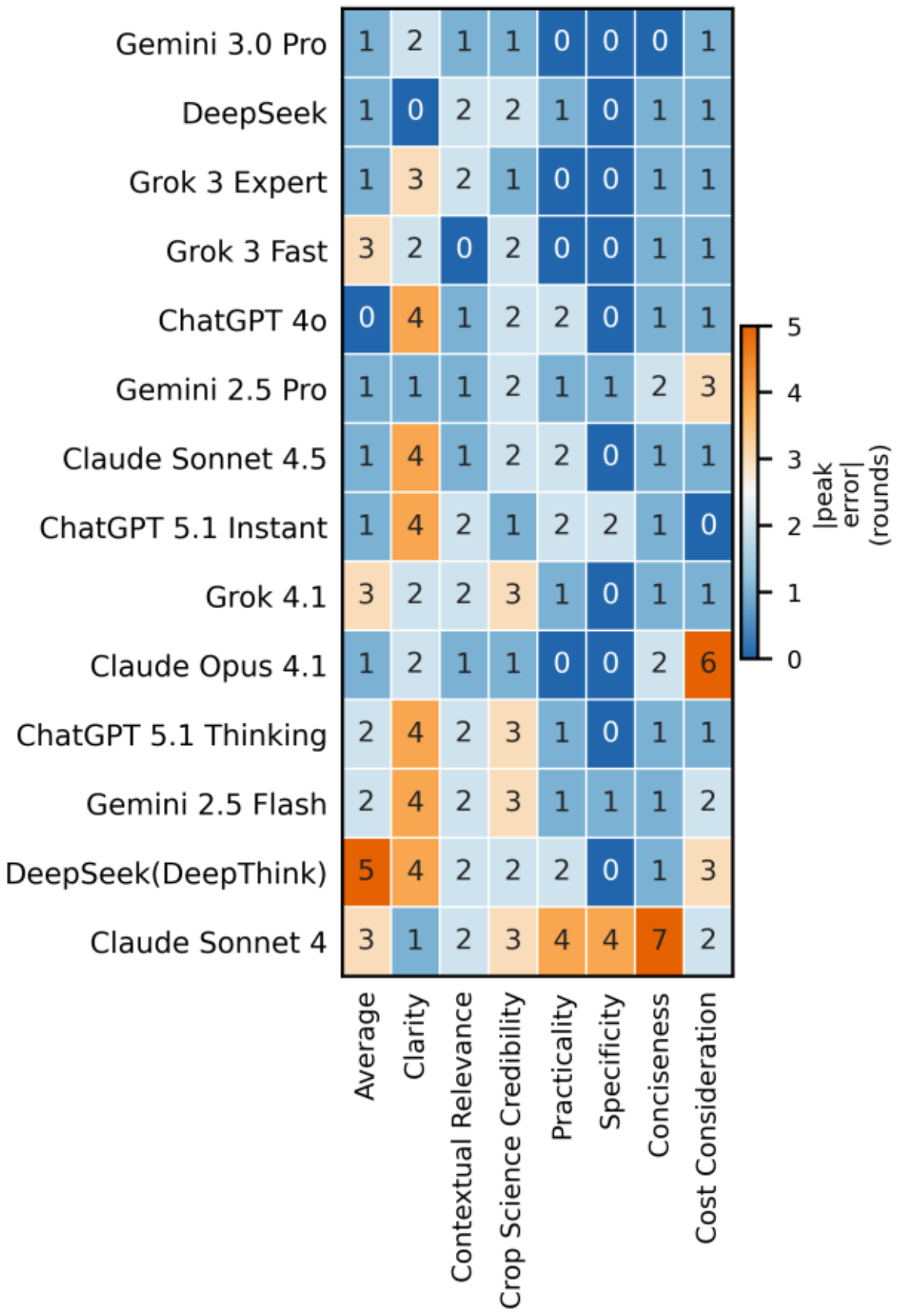

Fig. 6. Per-LLM peak-detection error matrix. Cell values give the absolute difference |LLM peak round − crop-scientist consensus peak round| for each criterion and the average score; rows are sorted by mean error. Blue cells (0–1) indicate that the judge's peak lies within one round of the expert consensus; orange cells (≥ 4) indicate large misses.

## 4. Discussion

### 4.1 Agentic XAI improved the recommendations in the early rounds

Our results demonstrate that the proposed agentic XAI framework substantially improved recommendation quality in the early refinement rounds. Both evaluator groups independently identified peak performance at Rounds 3–4, representing improvements of $\Delta = +1.23$ (crop scientists) and $\Delta = +1.44$ (LLMs) over Round 0 (Fig. 3; Table S1). This trajectory aligns with recent findings in iterative agentic workflows, where multi-turn reasoning loops rapidly converge to optimal solutions before plateauing (Madaan et al., 2023; Zhang et al., 2024).

By leveraging an LLM-driven agentic approach, our framework effectively bridges the persistent "interpretability–usability gap" in XAI applications (Hegselmann et al., 2023; Bilal et al., 2025), synthesizing raw SHAP-based outputs into intelligible narratives that facilitate practical decision-making for farmers (Wang et al., 2024). The convergent validation by both human domain experts and LLM-as-judge evaluators (Table 4) strengthens confidence in these quality improvements and indicates that the framework translates technical XAI outputs into actionable guidance.

### 4.2 Early stopping is necessary: the bias–variance trade-off

The improvement, however, was not monotonic, and our results support the hypothesized bias–variance analogy in explanation comprehensiveness. In both evaluator groups, average quality followed an inverted U-shaped trajectory identified by derivative-based GAM criteria, with an optimal range in the early-to-mid rounds (Rounds 2–4) followed by deterioration (Fig. 3; Tables S1 and S2). In this framing, early rounds lack analytical depth (analogous to bias), whereas continuing beyond the optimum introduces verbosity and abstraction that obscure actionable insights (analogous to variance).

The bias side appears in Rounds 0–2. At Round 0, the MLLM interpreted the SHAP beeswarm plot (Fig. 2) alone, without supplementary quantitative analyses, and its recommendations lacked the specificity and agronomic detail that actionable guidance requires (Fig. 4). The gains in six metrics (all but Cost Consideration) from Round 0 to Rounds 2–4 reflect the analytical depth that refinement added (Fig. 5; Table 3).

The variance side emerges from Round 5 onwards in several forms. Conciseness, which peaked early and then declined in both groups, captures the growing verbosity that overwhelms lay readers (Fig. 5). Beyond length, the MLLM drifted toward abstraction and generalization, producing recommendations that were analytically sophisticated (Table 3) but detached from implementable practice. Cost Consideration exemplifies this drift. Because the dataset contained no economic variables (Table 1), its steady improvement rested on economic reasoning generated without grounding data, precisely the elaborate but unfounded content that we hypothesized as "high variance" (Fig. 5). The late-round decline of Contextual Relevance, Crop Science Credibility, Practicality, and Specificity in both groups shows that the practical criteria are the most vulnerable to this degradation. They lost the concrete specificity that makes recommendations actionable even as analytical sophistication increased, and their divergence from Cost Consideration and Conciseness shows that the criteria cannot be optimized simultaneously by iteration alone.

This pattern aligns with recent findings on diminishing returns in iterative AI systems, where models lose substantial capabilities within limited refinement cycles because of reward overoptimization and distribution shift (Zhu et al., 2024; Rafailov et al., 2024; Adnan et al., 2025). Recent studies of LLM refinement point the same way. Without targeted feedback, most models fail to improve their answers consistently over successive turns (Lee et al., 2026), and in document-level translation the gains from refinement come mainly from fluency, with limited improvement in adequacy (Tan et al., 2026). In our agricultural context, refinement optimized analytical comprehensiveness and economic completeness while sacrificing the operational specificity and contextual grounding that farmer decisions require. Early stopping after 3–4 rounds therefore outperformed extended iteration, extending emerging evidence from machine learning research (Snell et al., 2024; Zhu et al., 2024) to knowledge translation, and it should be treated as a design principle of agentic XAI rather than a tuning detail.

### 4.3 Deciding when to stop: LLM judges, screened panels, and human oversight

If early stopping is necessary, the practical question is who detects the peak. The agreement analysis (Table 4) delimits what LLM judges can contribute. They cannot replace experts for absolute quality assessment, because every criterion was overscored by 1.4–2.3 points and an LLM-reported score therefore carries no calibrated meaning on the expert scale. What the LLM panel largely preserved was the experts' ordering of rounds, including the location of the quality peak, and detecting that peak is an ordering task that is invariant to any constant shift in scores. Monitoring refinement therefore imposes a much weaker requirement on an LLM judge than conventional LLM-as-judge applications assume, since ordinal fidelity suffices and score calibration is unnecessary. In principle, then, the loop can be stopped by an LLM judge, and the panel average did locate the expert peak within one round on every score series.

In practice, however, relying on a single judge is risky, because individual judges differed widely (Fig. 6). The best judge located the expert peak within ±1 round on 87.5% of the score series and the worst on 12.5%, and the worst judge was the model that had generated the recommendations, which echoes the self-preference bias reported for LLM-based evaluation (Ye et al., 2024) and argues against letting an agent grade its own refinement. Two safeguards follow. First, use several judges rather than one, and screen them against expert-scored data: collect expert ratings once for a small pilot (in our case, 11 rounds scored by 12 crop scientists), benchmark candidate judges against the expert peak, retain only those whose peak falls within ±1 round on the criteria of interest (Fig. 6), and exclude models that share a family with the generating agent. The screened judges should also score against the same explicit criteria as the experts, which in this study were the guiding questions of Table 2, because the criteria given to an LLM judge are critical to its reliability (Yamauchi et al., 2026). Second, keep humans in the loop at the decision point: during subsequent refinement runs, use the screened judges' round ranking as the stopping signal, halting once the running maximum stops improving, and route only the selected output to experts for final validation. Compared with a fixed iteration budget, this hybrid procedure respects the finding that quality trajectories differ across criteria (Fig. 5), and it reduces expert effort from rating every round to validating a single output.

Two caveats bound this guidance. Screening presupposes at least one expert-scored pilot, so the procedure economizes expert effort rather than eliminating it. And the procedure is operational guidance distilled from a single case study, not a validated stopping algorithm; formal guarantees for judge-driven stopping remain future work.

### 4.4 Keeping the agentic process transparent and accountable

Whoever decides when to stop, deploying agentic XAI raises risks beyond the choice of iteration count. From a principal–agent perspective, the relationship between farmers and agentic systems introduces information asymmetries and potential misalignment between system objectives and farmer interests (Gabison & Xian, 2025), and unclear liability frameworks may impede adoption when agent-generated advice leads to poor outcomes. Such concerns already surface in extension practice, where agents hesitate to rely fully on generative AI out of concern for professional judgment and accountability (Gouroubera et al., 2026), and where LLM tools, mostly developed by private firms, raise questions of transparency and accountability (Edet & Chowdhury, 2026). Our results expose a specific vulnerability: the agent has no internal mechanism for recognizing the groundlessness of its own reasoning. It continued to generate increasingly complex analyses, including economic reasoning unsupported by any economic data, while expert-judged quality deteriorated (Fig. 5; Table 3). Agentic generation and execution of analytical code add further hazards, namely unintended data manipulation, accumulation of analytical errors across iterations, and propagation of flawed analyses through successive rounds without external validation (Raza et al., 2025).

The refinement process itself also introduces opacity that may reduce rather than enhance trust, even though agentic XAI aims to improve interpretability. Non-deterministic execution flows and unpredictable behavioral variability create "black box" challenges that hinder accountability and comprehension (Moshkovich et al., 2025). Farmers must trust the original model, its SHAP interpretation, the LLM's self-generated analyses, the code producing those analyses, and the synthesis of multiple information sources, and this chain lengthens with every iteration. Even optimally stopped refinement therefore demands more trust than simpler advisory approaches.

Observability is the remedy that this study can offer. Systematic examination of intermediate outputs by domain experts establishes observability into the agentic refinement process (Limonad, 2025; Moshkovich et al., 2025). Archiving the agentically generated code, visualizations, and recommendations of every round, as done in the Zenodo repository of this study (Yamaguchi et al., 2026), lets experts verify each iteration's analytical artifacts, identify analytical errors, and check agronomic soundness. Transparency of the process is, in this sense, the precondition for the screened, human-in-the-loop stopping described above to be trustworthy.

### 4.5 Methodological considerations and limitations

First, this study is based on data from a single crop system (rice) under specific geographical conditions (Fukushima, Japan), which limits generalizability. Different agricultural systems may exhibit distinct patterns of response to iterative refinement, particularly those with different complexity levels, decision-making timescales, or cultural contexts. The framework's performance with perennial crops, livestock systems, or integrated farming approaches remains unexplored. However, the convergent patterns across independent evaluator groups and the theoretical alignment with broader machine learning findings on iterative refinement suggest that the early stopping principle may generalize beyond this specific context.

Second, the agentic XAI framework relied on a single generation model (Claude Sonnet 4), so the observed refinement trajectories may be specific to its architecture or training data. The 14 evaluator LLMs, although diverse, may also carry correlated biases. Models from the same provider or model family often share training corpora and alignment objectives such as reinforcement learning from human feedback, and may therefore share evaluation tendencies, so the consistency across evaluators may partly reflect shared artifacts rather than independent validation. The per-LLM heterogeneity shown in Fig. 6 reinforces this caution, as agreement with the expert panel varied widely across judges and the judge sharing its model family with the generating agent performed worst.

Third, the evaluation is based on recommendation quality assessed at the time of generation, without field validation of implementation outcomes over time. While the quality metrics employed align with established agricultural decision support system evaluation frameworks (Zhai et al., 2020; Htun et al., 2022; Helps et al., 2024), the ultimate test of effectiveness lies in farmer adoption, implementation fidelity, and agricultural outcomes. Field studies that implement recommendations generated at different refinement levels would test our findings against agricultural outcomes and may reveal quality criteria that our framework does not capture.

### 4.6 Future research directions

The generalizability of the bias–variance pattern and of the early stopping principle requires validation beyond a single rice system. Other crops, livestock systems, and environmental management contexts may exhibit different refinement dynamics depending on their complexity, decision timescales, and data availability, and tests in medical, financial, or environmental advisory settings would reveal whether the trade-off is a general property of knowledge translation systems (Holzinger et al., 2022; Wang et al., 2024). Within the evaluation loop, three questions stand out. The judge-driven stopping procedure proposed here needs formal validation and adaptive stopping criteria that respond to real-time quality signals. The balance between automated LLM monitoring for efficiency, domain-expert validation for credibility, and end-user evaluation for practical utility needs to be established in line with human-centered and responsible-AI principles (Holzinger et al., 2022; Shneiderman, 2020; Gardezi et al., 2022; Dara et al., 2022). And the ultimate validation must come from end users, here farmers as laypersons, who judge whether recommendations are understandable, actionable, and valuable for their decisions (Zhai et al., 2020; Htun et al., 2022; Gardezi et al., 2023).

Future agentic XAI systems could also push the early stopping limit outward by incorporating additional grounding data once the initial performance peak is reached. The degradation beyond Round 4 reflects the finite information available to the agent. Integration protocols such as the Model Context Protocol (MCP) (Anthropic, 2024) or retrieval-augmented generation (RAG) architectures (Lewis et al., 2020; Gao et al., 2023), which agricultural advisory systems have begun to use to ground their answers (Ravindran et al., 2026; Sawant et al., 2026), could give agents access to real-time agricultural databases, domain literature, or organizational knowledge when the initial refinement approaches are exhausted, so that each data augmentation supports a further cycle of improvement before a new plateau is reached.

## 5. Conclusion

This study asked whether an agentic XAI system that iteratively refines SHAP-based explanations produces better explanations, and answered it with a case study in rice management judged by 12 crop scientists and 14 LLM judges. The framework was useful: refinement raised expert-judged recommendation quality by about one third within three to four rounds.

The improvement was not monotonic. Beyond the peak, further iteration degraded quality below the unrefined baseline through verbosity and ungrounded abstraction, and the seven quality criteria followed divergent trajectories that no single iteration count can reconcile. Strategic early stopping is therefore a design principle of agentic XAI, not a tuning detail.

Deciding when to stop can be delegated in part to LLM judges, which preserved the experts' ranking of rounds well enough to flag the peak. Their inflated scores and their heterogeneity, including the failure of the generating model to judge its own output, mean that several judges screened against expert ratings and combined with human validation of the selected output are safer than any single judge. All of this presupposes a transparent process in which the code, figures, and recommendations of every round are archived and open to expert scrutiny. Validating these principles across crops and domains, formalizing judge-driven stopping, and testing recommendations with farmers themselves are the next steps.

**CRediT authorship contribution statement**

**Tomoaki Yamaguchi:** Conceptualization, Data curation, Formal analysis, Funding acquisition, Investigation, Methodology, Software, Resources, Validation, Visualization, Writing – original draft, Writing – review & editing. **Yutong Zhou:** Supervision, Writing – review & editing. **Masahiro Ryo:** Conceptualization, Funding acquisition, Investigation, Project administration, Validation, Writing – review & editing. **Keisuke Katsura:** Conceptualization, Funding acquisition, Investigation, Project administration, Supervision, Validation, Writing – review & editing.

**Funding**

This work was supported by the Ministry of Agriculture, Forestry and Fisheries of Japan through the commissioned project "Commissioned projects for promotion of strategic international joint research (Joint research with Germany)" [grant number JPJ008837]. The funder had no role in the study design, in the collection, analysis and interpretation of the data, in the writing of the report, or in the decision to submit the article for publication.

**Data availability**

The agentic-loop prompts and the code, figures, and recommendations generated in every refinement round are deposited in a single Zenodo record (https://doi.org/10.5281/zenodo.17876329; Yamaguchi et al., 2026), together with the analysis code, its outputs, and its software environment. The evaluation data from the 12 crop scientists and 14 LLM judges and the case-study farm data are available from the corresponding author on reasonable request.

**Declaration of competing interest**

The authors declare that they have no competing financial interests or personal relationships that may have influenced the work reported in this study.

**Declaration of generative AI and AI-assisted technologies in the manuscript preparation process**

During the preparation of this work, the authors used Claude Sonnet 4, Claude Sonnet 4.5, Claude Opus 4, Claude Opus 4.7, and Claude Opus 5 (Anthropic, United States) in order to assist with manuscript proofreading, literature organization, and manuscript restructuring and editing. After using this tool/service, the authors reviewed and edited the content as needed and take full responsibility for the content of the published article.

**Acknowledgements**

We express our sincere gratitude to the crop scientists who generously provided their time and expertise to evaluate the AI-generated recommendations in this study.

**References**

Acharya, D. B., Kuppan, K., & Divya, B. (2025). Agentic AI: Autonomous intelligence for complex goals—A comprehensive survey. *IEEE Access, 13*, 18912–18936. https://doi.org/10.1109/ACCESS.2025.3532853

Adnan, M., & Kuhn, C. C. N. (2025). The Debugging Decay Index: Rethinking debugging strategies for code LLMs. *arXiv preprint* arXiv:2506.18403. https://doi.org/10.48550/arXiv.2506.18403

Akaike, H. (1973). Information theory as an extension of the maximum likelihood principle. In B. N. Petrov & F. Csaki (Eds.), *Second International Symposium on Information Theory* (pp. 267–281). Akademiai Kiado.

Anthropic. (2024). *Introducing the Model Context Protocol*. https://www.anthropic.com/news/model-context-protocol

Bilal, A., Ebert, D., & Lin, B. (2025). LLMs for explainable AI: A comprehensive survey. *arXiv preprint* arXiv:2504.00125. https://doi.org/10.48550/arXiv.2504.00125

Burnham, K. P., & Anderson, D. R. (2002). *Model selection and multimodel inference: A practical information-theoretic approach* (2nd ed.). Springer. https://doi.org/10.1007/b97636

Chen, Z., Chen, H., Imani, M., Chen, R., & Imani, F. (2025). Vision language model for interpretable and fine-grained detection of safety compliance in diverse workplaces. Expert Systems with Applications, 265, 125769. https://doi.org/10.1016/j.eswa.2024.125769

Chern, S., Chern, E., Neubig, G., & Liu, P. (2024). Can large language models be trusted for evaluation? *arXiv preprint* arXiv:2401.16788. https://doi.org/10.48550/arXiv.2401.16788

Dara, R., Hazrati Fard, S. M., & Kaur, J. (2022). Recommendations for ethical and responsible use of artificial intelligence in digital agriculture. *Frontiers in Artificial Intelligence, 5*, Article 884192. https://doi.org/10.3389/frai.2022.884192

Davcev, D., Kalajdziski, S., Dimitrovski, I., Kitanovski, I., & Mitreski, K. (2026). Agentic AI-based IoT precision agriculture framework—Our vision and challenges. *AgriEngineering, 8*(4), Article 147. https://doi.org/10.3390/agriengineering8040147

De Clercq, D., Nehring, E., Mayne, H., & Mahdi, A. (2024). Large language models can help boost food production, but be mindful of their risks. *Frontiers in Artificial Intelligence, 7*, Article 1326153. https://doi.org/10.3389/frai.2024.1326153

Doshi-Velez, F., & Kim, B. (2017). Towards a rigorous science of interpretable machine learning. *arXiv preprint* arXiv:1702.08608. https://doi.org/10.48550/arXiv.1702.08608

Edet, U. I., & Chowdhury, A. (2026). Social acceptance of large language models in agricultural extension and advisory: A rapid review. *Q Open, 6*(1), Article qoag001. https://doi.org/10.1093/qopen/qoag001

Gabison, G. A., & Xian, R. P. (2025). Inherent and emergent liability issues in LLM-based agentic systems: A principal-agent perspective. *arXiv preprint* arXiv:2504.03255. https://doi.org/10.48550/arXiv.2504.03255

Gakhar, S., & Singh, R. K. (2026). Leveraging large language models (LLMs) for GeoAI-enabled digital agro-advisory. *Frontiers in Remote Sensing, 7*, Article 1839369. https://doi.org/10.3389/frsen.2026.1839369

Galaz, V., Schewenius, M., Donges, J. F., Fetzer, I., Zhivkoplias, E., Barfuss, W., Delannoy, L., Wang-Erlandsson, L., Gelbrecht, M., Heitzig, J., Hentati-Sundberg, J., Kennedy, C., Knecht, N., Lotcheris, R., Mahecha, M., Merrie, A., Montero, D., McPhearson, T., Mustafa, A., Nyström, M., Purves, D., Rocha, J. C., Ryo, M., van der Salm, C., Segun, S. T., Stephenson, A. B., Tellman, E., Tobar, F., & Vadrot, A. (2025). AI for a planet under pressure. *arXiv preprint* arXiv:2510.24373. https://doi.org/10.48550/arXiv.2510.24373

Gao, Y., Xiong, Y., Gao, X., Jia, K., Pan, J., Bi, Y., Dai, Y., Sun, J., Wang, M., & Wang, H. (2023). Retrieval-augmented generation for large language models: A survey. *arXiv preprint* arXiv:2312.10997. https://doi.org/10.48550/arXiv.2312.10997

Gardezi, M., Adereti, D. T., Stock, R., & Ogunyiola, A. (2022). In pursuit of responsible innovation for precision agriculture technologies. *Journal of Responsible Innovation, 9*(2), 224–247. https://doi.org/10.1080/23299460.2022.2071668

Gardezi, M., Joshi, B., Rizzo, D. M., Ryan, M., Prutzer, E., Brugler, S., & Dadkhah, A. (2023). Artificial intelligence in farming: Challenges and opportunities for building trust. *Agronomy Journal, 116*(3), 1217–1228. https://doi.org/10.1002/agj2.21353

Gouroubera, M. W., Segnon, A. C., Gouthon, M., Moumouni-Moussa, I., & Zougmoré, R. B. (2026). Generative AI use and advisory performance among agricultural extension agents in Benin. *Frontiers in Artificial Intelligence, 9*, Article 1763406. https://doi.org/10.3389/frai.2026.1763406

Hegselmann, S., Buendia, A., Lang, H., Agrawal, M., Jiang, X., & Sontag, D. (2023). TabLLM: Few-shot classification of tabular data with large language models. In *International Conference on Artificial Intelligence and Statistics (AISTATS)* (PMLR 206, pp. 5549–5581).

Heider, M., Albrecht, M., Schilp, J., & Hähner, J. (2026). Assessing stakeholder perspectives on the explainability of AI solutions for smart production planning with just-in-time logistics. Expert Systems with Applications, 314, 131352. https://doi.org/10.1016/j.eswa.2026.131352

Helps, J. C., van den Bosch, F., Paveley, N., Jørgensen, L. N., Holst, N., & Milne, A. E. (2024). A framework for evaluating the value of agricultural pest management decision support systems. *European Journal of Plant Pathology, 169*, 887–902. https://doi.org/10.1007/s10658-024-02878-1

Holzinger, A., Saranti, A., Angerschmid, A., Retzlaff, C. O., Gronauer, A., Pejakovic, V., Medel-Jimenez, F., Krexner, T., Gollob, C., & Stampfer, K. (2022). Digital transformation in smart farm and forest operations needs human-centered AI: Challenges and future directions. *Sensors, 22*(8), Article 3043. https://doi.org/10.3390/s22083043

Hosseini, S., & Seilani, H. (2025). The role of agentic AI in shaping a smart future: A systematic review. *Array*, Article 100399. https://doi.org/10.1016/j.array.2025.100399

Htun, N. N., Rojo, D., Ooge, J., De Croon, R., Kasimati, A., & Verbert, K. (2022). Developing visual-assisted decision support systems across diverse agricultural use cases. *Agriculture, 12*(7), Article 1027. https://doi.org/10.3390/agriculture12071027

Ibrahim, A., Senthilkumar, K., & Saito, K. (2024). Evaluating responses by ChatGPT to farmers' questions on irrigated lowland rice cultivation in Nigeria. *Scientific Reports, 14*(1), Article 3407. https://doi.org/10.1038/s41598-024-53916-1

Kaler, B., & Kaur, A. (2025). A systematic survey on explainable artificial intelligence (XAI) for plant health monitoring. *Applied Intelligence, 55*, Article 889. https://doi.org/10.1007/s10489-025-06790-x

Kim, S., Shin, J., Cho, Y., Jang, J., Longpre, S., Lee, H., Yun, S., Shin, S., Kim, S., Thorne, J., & Seo, M. (2024a). Prometheus: Inducing fine-grained evaluation capability in language models. *arXiv preprint* arXiv:2310.08491. https://doi.org/10.48550/arXiv.2310.08491

Kim, S., Suk, J., Longpre, S., Lin, B. Y., Shin, J., Welleck, S., Neubig, G., Lee, M., Lee, K., & Seo, M. (2024b). Prometheus 2: An open source language model specialized in evaluating other language models. In *Proceedings of the 2024 Conference on Empirical Methods in Natural Language Processing (EMNLP)* (pp. 4334–4353). https://doi.org/10.18653/v1/2024.emnlp-main.248

Kodaira, M., & Shibusawa, S. (2013). Using a mobile real-time soil visible-near infrared sensor for high resolution soil property mapping. *Geoderma, 199*, 64–79. https://doi.org/10.1016/j.geoderma.2012.09.007

Kumar, A., Zhuang, V., Agarwal, R., Su, Y., Co-Reyes, J. D., Singh, A., Baumli, K., Iqbal, S., Bishop, C., Roelofs, R., Zhang, L. M., McKinney, K., Shrivastava, D., Paduraru, C., Tucker, G., Precup, D., Behbahani, F., & Faust, A. (2024). Training language models to self-correct via reinforcement learning. *arXiv preprint* arXiv:2409.12917. https://doi.org/10.48550/arXiv.2409.12917

Lee, Y.-J., Kim, S., Lee, B.-K., Moon, M., Hwang, Y., Kim, J. M., Neubig, G., Welleck, S., & Choi, H.-J. (2026). RefineBench: Evaluating refinement capability of language models via checklists. In *International Conference on Learning Representations (ICLR) 2026*. https://openreview.net/forum?id=GYJFJz9Dy5

Lewis, P., Perez, E., Piktus, A., Petroni, F., Karpukhin, V., Goyal, N., Küttler, H., Lewis, M., Yih, W., Rocktäschel, T., Riedel, S., & Kiela, D. (2020). Retrieval-augmented generation for knowledge-intensive NLP tasks. In *Advances in Neural Information Processing Systems, 33* (pp. 9459–9474).

Li, Y., Wang, J., Yuan, Z., & Zhang, H. (2026). Agent technology for agricultural intelligence: Methodological framework and applications. *Electronics, 15*(8), Article 1547. https://doi.org/10.3390/electronics15081547

Limonad, L. (2025). Agentic AI process observability: Discovering behavioral variability. *arXiv preprint* arXiv:2505.20127. https://doi.org/10.48550/arXiv.2505.20127

Lin, L. I.-K. (1989). A concordance correlation coefficient to evaluate reproducibility. *Biometrics, 45*(1), 255–268. https://doi.org/10.2307/2532051

Lipton, Z. C. (2018). The mythos of model interpretability: In machine learning, the concept of interpretability is both important and slippery. *Queue, 16*(3), 31–57. https://doi.org/10.1145/3236386.3241340

Liu, Y., Iter, D., Xu, Y., Wang, S., Xu, R., & Zhu, C. (2023). G-Eval: NLG evaluation using GPT-4 with better human alignment. In *Proceedings of the 2023 Conference on Empirical Methods in Natural Language Processing (EMNLP)* (pp. 2511–2522). https://doi.org/10.18653/v1/2023.emnlp-main.153

Lundberg, S. M., & Lee, S. I. (2017). A unified approach to interpreting model predictions. In *Advances in Neural Information Processing Systems, 30* (pp. 4768–4777).

Madaan, A., Tandon, N., Gupta, P., Hallinan, S., Gao, L., Wiegreffe, S., Alon, U., Dziri, N., Prabhumoye, S., Yang, Y., Gupta, S., Majumder, B. P., Hermann, K., Welleck, S., Yazdanbakhsh, A., & Clark, P. (2023). Self-Refine: Iterative refinement with self-feedback. In *Advances in Neural Information Processing Systems, 36* (NeurIPS 2023).

Manafi Varkiani, S., Pattarin, F., Fabbri, T., & Fantoni, G. (2025). Predicting employee attrition and explaining its determinants. Expert Systems with Applications, 272, 126575. https://doi.org/10.1016/j.eswa.2025.126575

Molnar, C. (2020). *Interpretable machine learning*. https://christophm.github.io/interpretable-ml-book/

Moshkovich, D., Mulian, H., Zeltyn, S., Eder, N., Skarbovsky, I., & Abitbol, R. (2025). Beyond black-box benchmarking: Observability, analytics, and optimization of agentic systems. *arXiv preprint* arXiv:2503.06745. https://doi.org/10.48550/arXiv.2503.06745

Ohno, H., Sasaki, K., Ohara, G., & Nakazono, K. (2014). Development of grid square air temperature and precipitation data compiled from observed, forecasted, and climatic normal data. *Climate in Biosphere, 14*, 71–79. https://doi.org/10.2480/cib.J-16-028

Pang, R. Y., Yuan, W., Cho, K., He, H., Sukhbaatar, S., & Weston, J. (2024). Iterative reasoning preference optimization. In *Advances in Neural Information Processing Systems, 37* (NeurIPS 2024). https://doi.org/10.52202/079017-3702

Rafailov, R., Chittepu, Y., Park, R., Sikchi, H., Hejna, J., Knox, W. B., Finn, C., & Niekum, S. (2024). Scaling laws for reward model overoptimization in direct alignment algorithms. In *Advances in Neural Information Processing Systems, 37* (NeurIPS 2024, pp. 126207–126242). https://doi.org/10.52202/079017-4009

Raharimanana, V., Yamaguchi, T., Tsujimoto, Y., Oo, A. Z., Nishigaki, T., Rakotonindrina, H., & Katsura, K. (2023). A machine learning approach is effective to elucidate yield-limiting factors of irrigated lowland rice under heterogeneous growing conditions and management practices. *Field Crops Research, 304*, Article 109170. https://doi.org/10.1016/j.fcr.2023.109170

Rajbongshi, A., Johora, F. T., Hossain, A., Sarker, M. S., Rahman, M. H., Rahman, M. W., Alotaibi, F. T., & Moni, M. A. (2026). Leveraging explainable AI for sustainable agriculture: A comprehensive review of recent advances. *Artificial Intelligence Review, 59*(3), Article 105. https://doi.org/10.1007/s10462-025-11459-5

Ravindran, D. J. S., Skarga-Bandurova, I., V, S., Awais, M., & S, M. (2026). AgroLLM: Connecting farmers and agricultural practices through large language models for enhanced knowledge transfer and practical application. *AgriEngineering, 8*(1), Article 38. https://doi.org/10.3390/agriengineering8010038

Raza, S., Qureshi, R., Zahid, A., Kamawal, S., Sadak, F., Fioresi, J., Saeed, M., Sapkota, R., Jain, A., Zafar, A., Hassan, M. U., Zafar, A., Maqbool, H., Vayani, A., Wu, J., & Shoman, M. (2025). Who is responsible? The data, models, users or regulations? Responsible generative AI for a sustainable future. *arXiv preprint* arXiv:2502.08650. https://doi.org/10.48550/arXiv.2502.08650

Ribeiro, M. T., Singh, S., & Guestrin, C. (2016). "Why should I trust you?" Explaining the predictions of any classifier. In *Proceedings of the 22nd ACM SIGKDD International Conference on Knowledge Discovery and Data Mining* (pp. 1135–1144). https://doi.org/10.1145/2939672.2939778

Ryo, M. (2022). Explainable artificial intelligence and interpretable machine learning for agricultural data analysis. *Artificial Intelligence in Agriculture, 6*, 257–265. https://doi.org/10.1016/j.aiia.2022.11.003

Sapkota, R., Roumeliotis, K. I., & Karkee, M. (2025). AI agents vs. agentic AI: A conceptual taxonomy, applications and challenges. *arXiv preprint* arXiv:2505.10468. https://doi.org/10.48550/arXiv.2505.10468

Sawant, S., Nair, R., & Hariharan, S. (2026). Empowering farmers with artificial intelligence: A retrieval-augmented generation based large language model advisory framework. *Journal of Agricultural Engineering, 57*(2). https://doi.org/10.4081/jae.2026.1908

Schiller, J., Stiller, S., & Ryo, M. (2025). Artificial intelligence in environmental and Earth system sciences: Explainability and trustworthiness. *Artificial Intelligence Review, 58*, Article 316. https://doi.org/10.1007/s10462-025-11165-2

Shneiderman, B. (2020). Human-centered artificial intelligence: Reliable, safe & trustworthy. *International Journal of Human–Computer Interaction, 36*(6), 495–504. https://doi.org/10.1080/10447318.2020.1741118

Simpson, G. L. (2018). Modelling palaeoecological time series using generalised additive models. *Frontiers in Ecology and Evolution, 6*, Article 149. https://doi.org/10.3389/fevo.2018.00149

Singh, N., Wang'ombe, J., Okanga, N., Zelenska, T., Repishti, J., G K, J., Mishra, S., Manokaran, R., Singh, V., Rafiq, M. I., Gandhi, R., & Nambi, A. (2024). Farmer.Chat: Scaling AI-powered agricultural services for smallholder farmers. *arXiv preprint* arXiv:2409.08916. https://doi.org/10.48550/arXiv.2409.08916

Snell, C., Lee, J., Xu, K., & Kumar, A. (2024). Scaling LLM test-time compute optimally can be more effective than scaling model parameters. *arXiv preprint* arXiv:2408.03314. https://doi.org/10.48550/arXiv.2408.03314

Swati, N. L. P., Gupta, S. V., Duddela, N. S., & Parvathy, L. R. (2026). Agentic AI-driven autonomous decision support system for smart agriculture. *Scientific Reports, 16*(1), Article 9972. https://doi.org/10.1038/s41598-026-39472-w

Tan, S., Zhu, D., Tran, K., Denkowski, M., Trenous, S., Ribeiro, L. F. R., Byrne, B., & Hieber, F. (2026). What does LLM refinement actually improve? A systematic study on document-level literary translation. In *Proceedings of the 64th Annual Meeting of the Association for Computational Linguistics (Volume 1: Long Papers)* (pp. 5929–5957). https://doi.org/10.18653/v1/2026.acl-long.268

Tzachor, A., Devare, M., King, B., Avin, S., & Ó hÉigeartaigh, S. (2022). Responsible artificial intelligence in agriculture requires systemic understanding of risks and externalities. *Nature Machine Intelligence, 4*, 104–109. https://doi.org/10.1038/s42256-022-00440-4

von Brömssen, C., Betner, S., Fölster, J., & Eklöf, K. (2021). A toolbox for visualizing trends in large-scale environmental data. *Environmental Modelling & Software, 136*, Article 104949. https://doi.org/10.1016/j.envsoft.2020.104949

Wang, L., Ma, C., Feng, X., Zhang, Z., Yang, H., Zhang, J., Chen, Z., Tang, J., Chen, X., Lin, Y., Zhao, W. X., Wei, Z., & Wen, J. (2024). A survey on large language model based autonomous agents. *Frontiers of Computer Science, 18*, Article 186345. https://doi.org/10.1007/s11704-024-40231-1

Wei, J., Wang, X., Schuurmans, D., Bosma, M., Ichter, B., Xia, F., Chi, E., Le, Q. V., & Zhou, D. (2022). Chain-of-thought prompting elicits reasoning in large language models. In *Advances in Neural Information Processing Systems, 35* (pp. 24824–24837).

Wu, Q., Bansal, G., Zhang, J., Wu, Y., Li, B., Zhu, E., Jiang, L., Zhang, X., Zhang, S., Liu, J., Awadallah, A. H., White, R. W., Burger, D., & Wang, C. (2023). AutoGen: Enabling next-gen LLM applications via multi-agent conversation framework. *arXiv preprint* arXiv:2308.08155. https://doi.org/10.48550/arXiv.2308.08155

Yamaguchi, T., Takamura, T., Tanaka, T. S. T., Ookawa, T., & Katsura, K. (2025a). A study on optimal input images for rice yield prediction models using CNN with UAV imagery and its reasoning using explainable AI. *European Journal of Agronomy, 164*, Article 127512. https://doi.org/10.1016/j.eja.2025.127512

Yamaguchi, T., Angeles, O., Iizumi, T., Dobermann, A., Katsura, K., & Saito, K. (2025b). Machine learning reveals drivers of yield sustainability in five decades of continuous rice cropping. *Field Crops Research, 333*, Article 110114. https://doi.org/10.1016/j.fcr.2025.110114

[dataset] Yamaguchi, T., Zhou, Y., Ryo, M., & Katsura, K. (2026). *Supplementary materials for: Agentic explainable artificial intelligence (Agentic XAI) to explore better explanations: A case study in decision support for rice cultivation in Japan* [Data set]. Zenodo. https://doi.org/10.5281/zenodo.17876329

Yamauchi, Y., Yano, T., & Oyamada, M. (2026). An empirical study of LLM-as-a-judge: How design choices impact evaluation reliability. In *Proceedings of the Fifth Workshop on Generation, Evaluation and Metrics (GEM)* (pp. 167–176). https://doi.org/10.18653/v1/2026.gem-main.19

Yao, S., Zhao, J., Yu, D., Du, N., Shafran, I., Narasimhan, K., & Cao, Y. (2023). ReAct: Synergizing reasoning and acting in language models. In *International Conference on Learning Representations (ICLR) 2023*.

Ye, J., Wang, Y., Huang, Y., Chen, D., Zhang, Q., Moniz, N., Gao, T., Geyer, W., Huang, C., Chen, P., Chawla, N. V., & Zhang, X. (2024). Justice or prejudice? Quantifying biases in LLM-as-a-Judge. *arXiv preprint* arXiv:2410.02736. https://doi.org/10.48550/arXiv.2410.02736

Zhai, Z., Martínez, J. F., Beltran, V., & Martínez, N. L. (2020). Decision support systems for agriculture 4.0: Survey and challenges. *Computers and Electronics in Agriculture, 170*, Article 105256. https://doi.org/10.1016/j.compag.2020.105256

Zhang, J., Hou, Y., Xie, R., Sun, W., McAuley, J., Zhao, W. X., Lin, L., & Wen, J. R. (2024). AgentCF: Collaborative learning with autonomous language agents for recommender systems. In *Proceedings of the ACM Web Conference 2024 (WWW '24)* (pp. 3679–3689). https://doi.org/10.1145/3589334.3645537

Zheng, L., Chiang, W. L., Sheng, Y., Zhuang, S., Wu, Z., Zhuang, Y., Lin, Z., Li, Z., Li, D., Xing, E., Zhang, H., Gonzalez, J. E., & Stoica, I. (2023). Judging LLM-as-a-Judge with MT-Bench and Chatbot Arena. In *Advances in Neural Information Processing Systems, 36* (NeurIPS 2023).

Zhu, B., Jordan, M., & Jiao, J. (2024). Iterative data smoothing: Mitigating reward overfitting and overoptimization in RLHF. In *Proceedings of the 41st International Conference on Machine Learning (ICML)* (PMLR 235, pp. 62405–62428).